\documentclass[11pt]{article}

\usepackage[preprint]{acl}

\usepackage{times}
\usepackage{latexsym}

\usepackage[T1]{fontenc}

\usepackage[utf8]{inputenc}

\usepackage{microtype}

\usepackage{inconsolata}

\usepackage{graphicx}

\usepackage{hyperref}       
\usepackage{url}            
\usepackage{booktabs}       
\usepackage{amsfonts}       
\usepackage{nicefrac}       
\usepackage{microtype}      
\usepackage{xcolor}         
\definecolor{webblue}{RGB}{93, 169, 221}
\usepackage{pifont}

\usepackage{svg}
\usepackage{graphicx}
\usepackage{multirow}
\usepackage{fontawesome5}

\usepackage[most]{tcolorbox}
\usepackage{tcolorbox}
\tcbuselibrary{breakable,skins}
\usepackage{dblfloatfix} 
\tcbset{
  guiPrompt/.style={
    colback=gray!10,            
    colframe=black,             
    coltitle=white,             
    colbacktitle=gray!30!black, 
    fonttitle=\bfseries,
    fontupper=\ttfamily,        
    boxrule=0.5pt,
    arc=2mm,                    
    top=0.6em, bottom=0.6em, left=0.8em, right=0.8em,
    enhanced,
    breakable
  }
}

\usepackage{custom_commands}
\usepackage{cleveref}

\title{PRO-CUA: Process-Reward Optimization for Computer Use Agents}

\author{Yifei He \quad
  Rui Yang \quad 
  Hao Bai \quad
  Tong Zhang \quad
  Han Zhao\\
  University of Illinois Urbana-Champaign \\
 [0.5em]
  \makebox[\textwidth][c]{
    \href{https://yifei-he.github.io/pro-cua-website/}{%
        \raisebox{-0.1\height}{\faGlobe} 
        \texttt{\textcolor{webblue}{Website}}%
    }
    \quad 
    \href{https://github.com/yifei-he/PRO-CUA}{%
        \raisebox{-0.1\height}{\faGithub} 
        \texttt{\textcolor{webblue}{Code}}%
    }
    \quad  
    \href{https://huggingface.co/PRO-CUA}{%
    \raisebox{-0.25\height}{\includegraphics[height=1.4em]{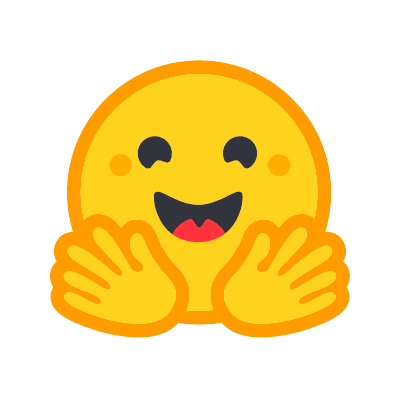}} 
    \texttt{\textcolor{webblue}{Model}}%
}
  }}

\begin{document}
\maketitle
\begin{abstract}
Computer use agents (CUAs) have shown strong potential for automating complex digital workflows, yet their training remains constrained by costly live environment interaction and limited high-quality supervision. Existing filtered behavior cloning pipelines suffer from imitation bottlenecks, including distribution shift from the expert demonstration and the absence of negative learning signals. Meanwhile, standard trajectory-level reinforcement learning struggles with sparse rewards, ambiguous credit assignment, and high infrastructure costs for long-horizon GUI interaction. In this work, we propose PRO-CUA, a process-reward optimization framework for training CUAs with iterative step-level reinforcement learning. PRO-CUA decouples on-policy environment interaction from policy optimization: the current policy collects states through live rollouts, generates diverse candidate actions for each state, receives step-level feedback from a process reward model (PRM), and is optimized with group-relative advantages. This design enables dense and flexible credit assignment without relying on golden answers or offline expert trajectories, while reducing distribution shift by training on the agent's own execution states. Experiments on live web benchmarks demonstrate the effectiveness of PRO-CUA and the reliability of PRM-guided step-level training.
\end{abstract}

\section{Introduction}
Driven by rapid breakthroughs in multimodal reasoning, autonomous agents are evolving into economically valuable digital coworkers. Computer use agents (CUAs)~\citep{operator,Claude-computer-use,agasheagent,qin2025ui,wang2025ui,wang2025opencua} have proven highly capable of seamlessly automating complex, open-ended workflows. By perceiving visual interfaces and executing sequential plans, these agents natively operate across diverse digital ecosystems, including web browsers~\citep{deng2023mind2web,he2024webvoyager,xue2025illusion} and desktop environments~\citep{xie2024osworld,bonattiwindows,wu2025gui}. Despite their immense commercial value and recent capability leaps, creating reliably generalized CUAs remains fundamentally bottlenecked by how they are trained. Specifically, researchers are constrained by two interlocking challenges: the prohibitive latency and computational cost of interacting with live GUI environments, and an acute scarcity of high-quality training data.

The most intuitive and prevailing approach to training CUAs is filtered behavior cloning (FBC) from expert demonstrations~\citep{bai2024digirl,xuaguvis,he2024openwebvoyager,shen2025thinking,he2026webstarscalabledatasynthesis}. However, FBC inherently suffers from imitation bottlenecks: it over-penalizes reasoning diversity, disproportionately overfits to easy tasks, and lacks negative learning signals for agents to learn from mistakes. To overcome these limitations, reinforcement learning (RL) offers a principled alternative. Yet, applying standard, \textit{trajectory-level} RL to CUAs introduces optimization and infrastructural difficulties. In long-horizon computer use tasks, receiving a single sparse reward at task completion makes step-wise credit assignment highly ambiguous, so the agent cannot deduce which specific action among dozens caused the ultimate failure. Furthermore, standard synchronous RL frameworks such as verl~\citep{sheng2025hybridflow} are computationally ill-equipped for multi-turn agent workflows. The latency of live GUI execution combined with the compounding memory costs of token-heavy image contexts renders trajectory-level optimization especially challenging. \looseness=-1

\begin{figure*}[t!]
    \centering
    \includegraphics[width=0.9\linewidth]{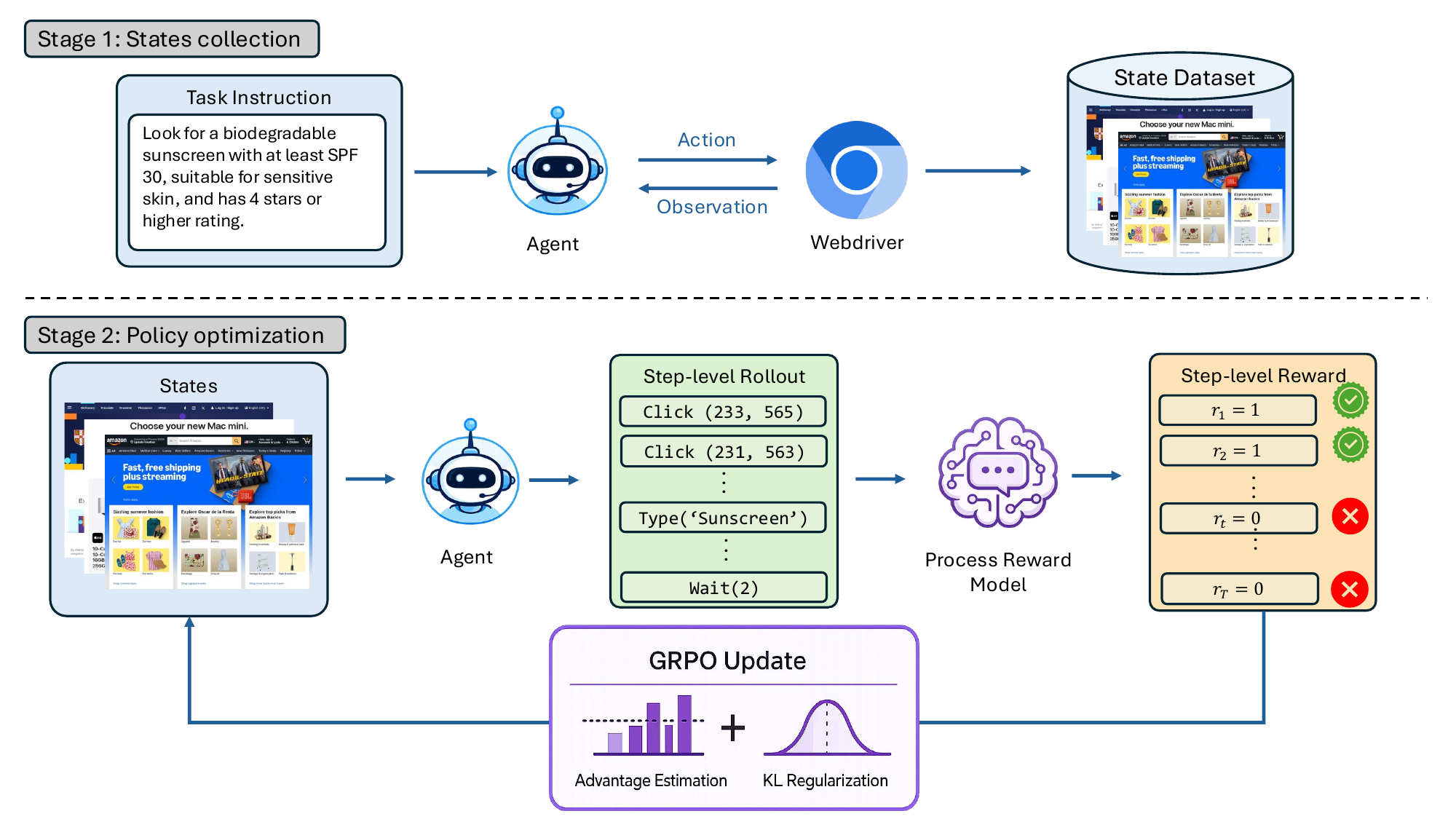}
    \caption{\textbf{Overview of the PRO-CUA pipeline.} PRO-CUA alternates between two stages across multiple training iterations. In \textbf{Stage 1}, the current policy interacts with the live environment to collect on-policy states. In \textbf{Stage 2}, policy optimization is performed without further environment interaction through three steps: \textbf{i) Step-level generation:} The agent samples multiple candidate actions for each collected state; \textbf{ii) PRM grading:} A process reward model assigns binary step-level rewards; and \textbf{iii) GRPO update:} The policy is optimized using group-relative advantages. The updated policy is then used for the next round of on-policy state collection.}
    \label{fig:main}
    \vspace{-0.4cm}
\end{figure*}

To bypass the sparse reward problem of trajectory-level learning, recent efforts have shifted toward \textit{step-level} RL paradigms~\citep{luo2025gui,yang2026gui}. While step-wise optimization makes credit assignment tractable, existing methods remain constrained by their reward design and data collection mechanisms. They predominantly rely on \textit{rule-based rewards }that require exact-match accuracy against a \textit{golden answer}, a strict requirement that drastically limits the scalability of viable training data. Moreover, these pipelines are primarily \textit{off-policy}, optimizing over states collected by a stronger teacher model rather than the target policy itself. Because an agent's actions sequentially alter future GUI observations, this off-policy collection introduces compounding distribution shift~\citep{ross2011reduction}. The offline training data can diverge from the actual, suboptimal states the agent will encounter, resulting in brittle performance and an inability to recover from mistakes.

To tackle the dual challenges of reward scarcity and off-policy distribution shift, we propose an on-policy self-evolvement framework termed PRO-CUA (Process-Reward Optimization for Computer Use Agents). As shown in \Cref{fig:main}, PRO-CUA alternates between two stages across multiple training iterations. In Stage 1, the current policy interacts with the live environment at an elevated sampling temperature to collect on-policy states. In Stage 2, policy optimization is performed without further environment interaction: for each collected state, the agent samples $G$ diverse thought-action pairs, a Process Reward Model (PRM) assigns binary step-level rewards, and the policy is updated with GRPO. This design decouples live environment interaction from policy optimization, allowing each stage to run with infrastructure tailored to its own computational profile while training the agent on its own execution states. \looseness=-1

Empirically, we verify the effectiveness of the PRO-CUA pipeline on live web benchmarks, including WebVoyager~\citep{he2024webvoyager}, Mind2Web-Live~\citep{pan2024webcanvas} and Online Mind2Web~\citep{xue2025illusion}. We further demonstrate the reliability of our PRM grading pipeline and the effective data utilization of our training approach.

Overall, our PRO-CUA framework provides a scalable pipeline for training computer use agents. In summary, our main contributions are: \textbf{i) Tailored CUA infrastructure design:} We decouple environment interaction and  model training, avoiding the system challenges to simultaneously perform agent rollout, environment interaction and policy optimization. \textbf{ii) On-policy state collection:} We eliminate the reliance on offline expert demonstrations, and enable the agent to learn from its own execution distribution. \textbf{iii) Dense and flexible credit assignment via PRMs:} We transition from sparse trajectory-level rewards to PRM-graded step-level GRPO. This provides fine-grained supervision without the requirement of collecting golden answers from expert demonstration. \looseness=-1


\section{Preliminaries}

\paragraph{Computer use agents (CUAs)} CUAs take sequential steps to interact with a graphical user interface (GUI) to complete a task defined in the task instruction. Following the ReAct~\citep{yaoreact} paradigm, agents typically generate interleaved thoughts and actions, explicitly externalizing their reasoning process to improve task execution. At any given step $n$, the agent receives a state context $x_n$, which comprises the instruction $\mathcal{I}$, the historical sequence of past thoughts and actions $\{(t_{i},a_{i})\}_{i=1}^{n-1}$, and a truncated window of the $w$ most recent visual observations (screenshots) $\{o_{n-j}\}_{j=0}^{w-1}$. Truncation is applied exclusively to the visual inputs to manage the prohibitive token costs associated with multimodal contexts. In this work, we set $w=1$, so the agent observes only the most recent screenshot, which is memory-efficient and has been shown to be sufficient for GUI perception~\citep{qin2025ui}. The agent's objective is to learn a policy $\pi_\theta(t_n, a_n \mid x_n)$ that iteratively selects actions leading to successful task completion. 

\paragraph{Filtered Behavior Cloning (FBC)} Currently, the most prevalent training paradigm for CUAs is FBC, which is conceptually equivalent to Rejection Sampling Fine-Tuning (RFT). In this approach, a dataset of candidate trajectories is filtered to retain only those that result in a successful final outcome, forming a curated dataset $\mathcal{D}_{\text{succ}}$. The policy is then optimized via standard supervised fine-tuning (SFT) to maximize the log-likelihood of the expert thoughts and actions:
\vspace{-0.3cm}
\begin{align*}
\mathcal{L}_{\text{SFT}}(\theta) = - \mathbb{E}_{\tau \sim \mathcal{D}_{\text{succ}}} \left[ \sum_{n=1}^{|\tau|} \log \pi_\theta(t_n, a_n \mid x_n) \right].
\end{align*}

\paragraph{Reinforcement Learning (RL)} While FBC optimizes the likelihood of both the thought tokens and action tokens, RL only requires a reward signal for the generated action. This distinction is important for CUAs: intermediate thoughts can be long and diverse, whereas task progress is ultimately determined by the executed action. RL therefore avoids forcing the policy to reproduce the reference reasoning trace, and instead reinforces any generation that leads to a rewarded action. \looseness=-1

In single-turn domains like mathematical reasoning, GRPO relies only on a sparse outcome reward (e.g., verifying the answer). However, this trajectory-level approach is often suboptimal for long-horizon computer use tasks due to credit assignment challenges. To address this issue, recent works deploy \textit{step-level} RL. Similar to SFT, given step context $x=\{\cI, \{t_{n-i}, a_{n-i}\}_{i=1}^{n-1}, \{o_{n-i}\}_{i=1}^w\}$, GRPO samples a group of $G$ candidate thought-action pairs $\{t_k,a_k\}_{k=1}^G\sim\pi_\theta(\cdot|x)$ and compute the reward $r_k=\cR(x,a_k)$, where $R$ is predominantly a rule-based reward based on a comparison with \textit{the golden answer} obtained through expert demonstration, which measures accuracy of the action based on the action type, the input text and the coordinates the agent interacts with (detailed formulation in Appendix \ref{appendix:rule_based_reward}). The overall objective is \looseness=-1
\begin{equation}\label{eq:grpo}
{\small
\begin{split}
&\mathcal{L}_{\text{GRPO}}(\theta) = -\mathbb{E}_{\substack{x \sim D \\ \{a_k\}_{k=1}^G \sim \pi_{\theta_{\text{old}}}(\cdot|x)}} \Bigg[ \frac{1}{G} \sum_{k=1}^G \min \Big( \rho_k \hat{A}_k, \\
&\quad \text{clip}(\rho_k, 1 - \epsilon, 1 + \epsilon) \hat{A}_k \Big) - \beta \cdot \text{KL}(\pi_\theta(\cdot|x) \,\|\, \pi_{\text{ref}}(\cdot|x)) \Bigg]
\end{split}
}
\end{equation}
where $\rho_k = \frac{\pi_\theta(a_k \mid x)}{\pi_{\theta_{\text{old}}}(a_k \mid x)}$ represents the importance sampling ratio, $\hat{A}_k$ is the relative advantage computed within the sampled group, and $\beta$ controls the KL divergence penalty against the reference model $\pi_{\text{ref}}$ to prevent policy collapse.

\section{PRO-CUA}
\subsection{On-policy State Collection}
Prior works primarily rely on distillation from strong teacher models or human demonstrations~\citep{he2026webstarscalabledatasynthesis,yang2026gui,luo2025gui}. However, these expert trajectories suffer from severe distribution shift: they fail to represent the sub-optimal states a developing agent actually encounters. Because an agent's primitive actions alter subsequent observations, it inevitably drifts into catastrophic or out-of-distribution states (e.g., stuck websites) that are entirely absent from expert data. \looseness=-1

While online RL offers a principled solution by allowing the agent to explore via its own policy, it introduces severe infrastructural bottlenecks in the CUA setting. Synchronizing high-throughput LLM inference (e.g., vLLM~\citep{kwon2023efficient}), high-latency web browser interactions, and dedicated training frameworks (e.g., verl~\citep{sheng2025hybridflow}) into a single loop results in prohibitive latency, I/O overhead, and hardware idling.

To overcome these bottlenecks while preserving the necessity of on-policy exploration, we propose a decoupled state collection paradigm. As illustrated in Stage 1 of \Cref{fig:main}, we disentangle the slow environment interactions from the compute-heavy policy optimization loop. We deploy the current agent policy to interact with live environments at an elevated sampling temperature, encouraging the discovery of diverse paths. During the agent rollout, we continuously harvest these exploratory trajectories, logging the task instructions, visual observations, and action histories into a state dataset $\mathcal{D}_{\text{state}}$. Formally, let $\mathcal{T}$ represent a set of collected trajectories (which may include both successful and failed rollouts). A single trajectory $\tau \in \mathcal{T}$ of length $|\tau|$ is composed of a sequence of step-level interactions. At each step $n \in \{1, \dots, |\tau|\}$, the environment and the agent's history dictate a state context $x_{n}^{(\tau)}$, formally defined as the tuple:
\vspace{-0.3cm}
\begin{align*}
\mathcal{D}_{\text{state}} = \bigcup_{\tau \in \mathcal{T}} \bigcup_{n=1}^{|\tau|} \{ x_{n}^{(\tau)} \},
\end{align*}
where $x_{n}^{(\tau)} = \left(\mathcal{I}^{(\tau)}, \{(t_{i},a_{i})\}_{i=1}^{n-1}, \{o_{n-j}\}_{j=0}^{w-1}\right)$. The dataset will be further employed in the next stage to optimize the policy.

This decoupled mechanism solves two critical problems. First, it allows for highly parallelized, scalable data generation without bottlenecking the training hardware. Second, by populating the dataset with the agent's own exploratory rollouts rather than expert demonstrations, the policy is forced to confront its own mistakes, natively generating the failure states required to learn error-recovery during the subsequent optimization stage.

\subsection{Step-level Rollout}
Given the on-policy state dataset collected in the previous stage, PRO-CUA proceeds to the step-level rollout process. In each training iteration, the agent samples states from $\mathcal{D}_{\text{state}}$ and, for each state, generates $G$ candidate thought-action pairs. These candidates are newly sampled from the current policy and are not the actions originally executed during state collection. Thus, the collected trajectories serve only to provide the state distribution encountered by the agent, rather than reference demonstrations for imitation. \looseness=-1

Crucially, the candidate actions are \textit{not executed in the live environment}. This avoids repeatedly invoking costly browser interactions inside the optimization loop, but also means that the resulting state transitions are not directly observed. We therefore defer evaluation to the PRM, as described in the next section. \looseness=-1


\subsection{Process Reward Model Grading}\label{sec:prm_grade}

\begin{figure}[t!]
    \centering
    \includegraphics[width=\linewidth]{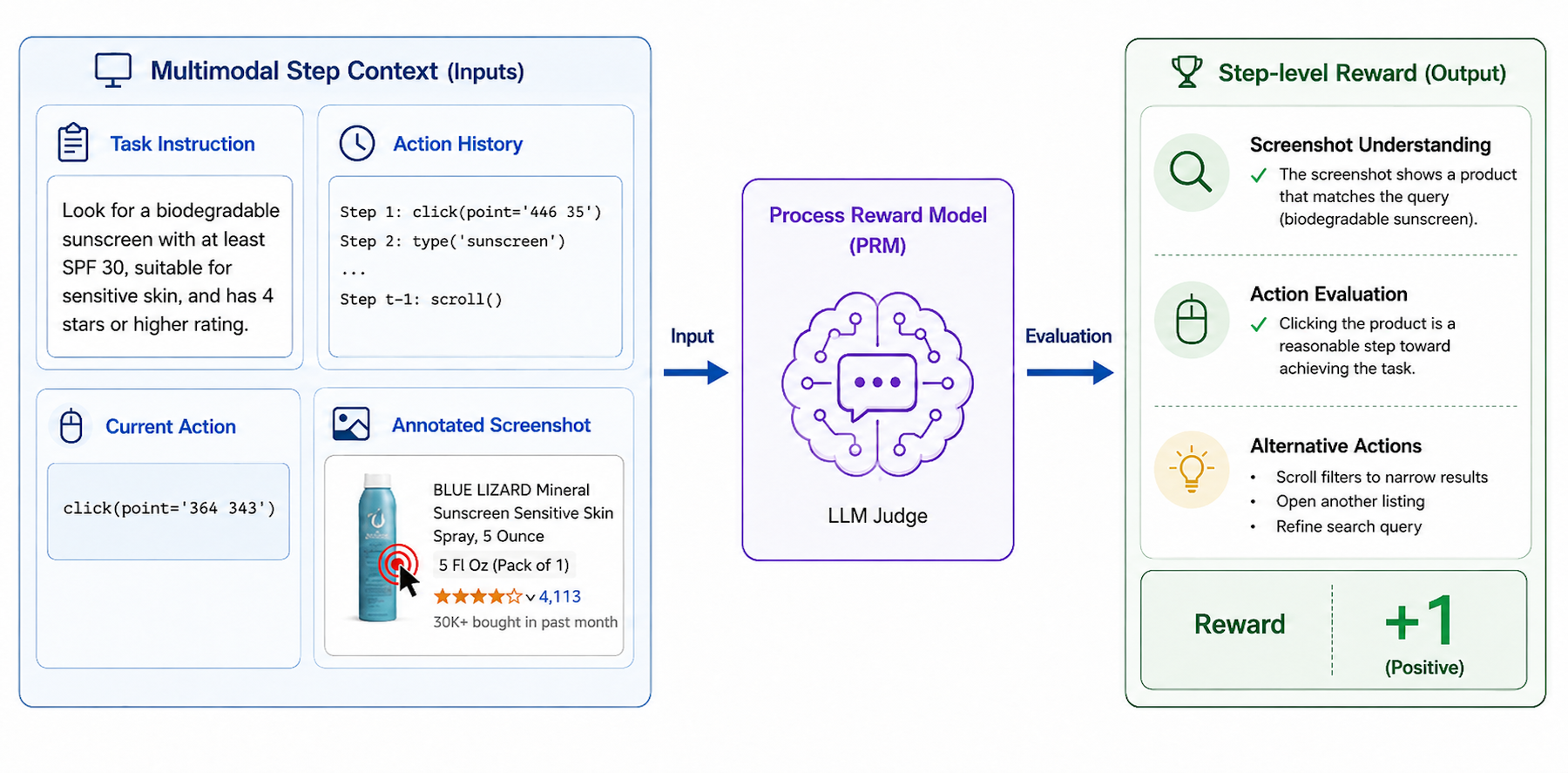}
    \caption{\textbf{Process Reward Model (PRM) grading pipeline.} The PRM receives a multimodal step context comprising the task instruction, the agent's action history, the proposed current action, and an annotated screenshot. For readability, the figure shows a zoomed-in crop, while the actual PRM input contains the full web interface. Based on this augmented context, the PRM generates a reasoning trace to assess whether the proposed action functionally advances the task, culminating in a binary step-level reward.}
    \label{fig:prm}
    \vspace{-0.3cm}
\end{figure}

A primary limitation of rule-based step rewards is their dependence on golden reference actions. Such rewards assign positive feedback only when the proposed action matches an expert action in action type, target element, and textual input. Consequently, they \textit{can only be applied to states from successful trajectories} where verified reference actions are available, and may penalize valid alternative actions different from the demonstration. This makes rule-based rewards data-inefficient and prevents the agent from learning from failure states.

PRO-CUA replaces these static heuristics with PRM grading. Instead of checking exact agreement with a predefined action, the PRM evaluates whether a proposed action functionally advances the task under the current state. This decouples step-level supervision from expert demonstrations and allows both successful and failed on-policy trajectories to contribute training states.

As illustrated in \Cref{fig:prm}, we formulate the PRM evaluation as a multimodal reasoning task. For any given step, the PRM is provided with a comprehensive context: the task instruction, the action history, and the agent's proposed current action. Since the candidate action is not executed in the live environment, the PRM does not observe the next state. To ground the evaluation, we annotate the screenshot at the target coordinates of the proposed action, allowing the PRM to identify the intended UI element. The PRM then reasons about whether the proposed action is visually grounded, non-redundant, and useful for task progress, and outputs a binary step-level reward. The detailed grading prompt is in Appendix \ref{appendix:prompt}.

Because the PRM is used as a training signal rather than an inference-time controller, PRO-CUA does not require perfectly calibrated rewards. GRPO distinguishes acceptable from unacceptable candidates within a sampled group, and the resulting updates aggregate reward signals across many states, sampled actions, and optimization steps. This makes the training process more tolerant to PRM noise than conventional test-time Best-of-N selection~\citep{chae2026webshepherd,zhang2026webarbiter}, where a single incorrect PRM preference can directly determine the executed action.

\subsection{Policy Optimization}
We optimize the policy using the step-level GRPO objective (\Cref{eq:grpo}). Unlike Filtered Behavior Cloning (FBC), which relies on exactly following expert demonstrations, GRPO offers two advantages. First, by relying on the PRM for functional evaluation rather than strict imitation, it preserves reasoning diversity, allowing the agent to freely explore novel thought paths as long as the resulting action is rewarded. Second, the group relative advantage naturally acts as a dynamic curriculum. Because advantages are mean-centered within the sampled group, gradients for trivial states wash out to zero, while difficult states (where most sampled actions fail) generate disproportionately high positive advantages for successful rollouts, focusing optimization where the policy struggles most.

\section{Experiments}
\subsection{Setup}

\paragraph{Rollout} We use \texttt{Qwen3-VL-4B-Instruct} and \texttt{Qwen3-VL-8B-Instruct}~\citep{bai2025qwen3} as base models. Following \citet{shen2025thinking}, we use their synthetic task set derived from WebVoyager. At each training iteration, we sample 256 tasks and roll out the current policy to collect on-policy trajectories, running 10 iterations in total. During rollout, we cap the maximum trajectory length at 20 steps, as most tasks are designed to be completed within this horizon, and use a temperature of 1.0 to encourage exploration.

\paragraph{Baselines} We compare PRO-CUA with two baseline training algorithms. Both baselines follow the same iterative setting: at each iteration, the current policy interacts with the environment to collect trajectories and obtain task-level outcome rewards. The collected data is then filtered or converted into training examples as follows: \textbf{Filtered Behavior Cloning (FBC)} keeps only successful trajectories and trains the policy on their observed thought-action sequences using SFT. \textbf{Step-level RL with rule-based rewards} retains states from successful trajectories and treats the corresponding executed actions as golden references. For each retained state, it samples $G$ candidate thought-action pairs from the current policy, assigns rule-based rewards by comparing each candidate action with the reference action, and optimizes the policy with GRPO. In both baselines, failed trajectories are discarded and do not contribute to training.

\paragraph{Training} For FBC experiments, we use LlamaFactory~\citep{zheng2024llamafactory} for training with a learning rate of 1e-5 for 2 epochs. For RL experiments, we use verl~\citep{sheng2025hybridflow} with a learning rate of 5e-6 for 1 epoch. All experiments are conducted on NVIDIA A6000 GPUs with 48G memory. We use a constant learning rate for RL training. Unlike standard offline training, our setting alternates between live environment rollouts and policy optimization, and the number of optimization steps in each iteration depends on the dynamically collected trajectories and their lengths. Therefore, the total number of updates is not known before rollout, making a pre-specified cosine decay schedule less natural. We instead use a constant learning rate for all RL methods to ensure a simple and consistent comparison across iterations.

\paragraph{Evaluation} We evaluate PRO-CUA on three online web benchmarks: WebVoyager~\citep{he2024webvoyager}, Mind2Web-Live~\citep{pan2024webcanvas}, and Online Mind2Web~\citep{xue2025illusion}. For each task, we allow the agent to take up to 30 steps. Due to website updates, some domains have introduced anti-scraping mechanisms, anti-bot checks, or access restrictions that were not present during benchmark construction. Following common practice for live web evaluation, we exclude domains that cannot be reliably accessed by automated agents, with details provided in Appendix~\ref{appendix:eval}. Following \citet{he2024webvoyager}, we provide the full agent trajectory and corresponding screenshots to GPT-5~\citep{singh2025openai}, which serves as an automatic evaluator for task success. \looseness=-1

\subsection{Reliability of the PRM Reward Signal}

\begin{table}[t!]
\centering
\caption{\textbf{Comparison of reward sources for step-level RL.} Under the same successful-trajectory training subset, PRM-based rewards outperform rule-based rewards, suggesting that visually grounded PRM feedback provides useful supervision for policy optimization.} \label{tab:reward}
\scalebox{0.8}{
\begin{tabular}{llr}
\toprule
Method     & Reward Type         & Success Rate \\ \midrule
Base model & / & 27.5         \\
Step-RL    & Rule-based          & 34.7         \\
Step-RL    & \texttt{Qwen3-VL-4B} PRM     & \textbf{36.6}        \\
Step-RL    & \texttt{GPT5-mini} PRM       & \textbf{36.8}        \\ \bottomrule
\end{tabular}}
\end{table}

A key question for PRO-CUA is whether PRM feedback provides a useful training signal compared with traditional rule-based rewards. To isolate the effect of the reward source, we conduct a controlled ablation on WebVoyager where all methods train on the same subset of states from successful trajectories. This setting favors rule-based rewards, since golden reference actions are available and failed trajectories are excluded. We compare rule-based rewards with two visually grounded PRMs, \texttt{Qwen3-VL-4B} and \texttt{GPT-5-mini}. As shown in \Cref{tab:reward}, both PRM-based variants outperform the rule-based baseline, indicating that PRM feedback can provide effective step-level supervision even under the same data regime as rule-based rewards.

\begin{figure}[t!]
    \centering
    \includegraphics[width=0.8\linewidth]{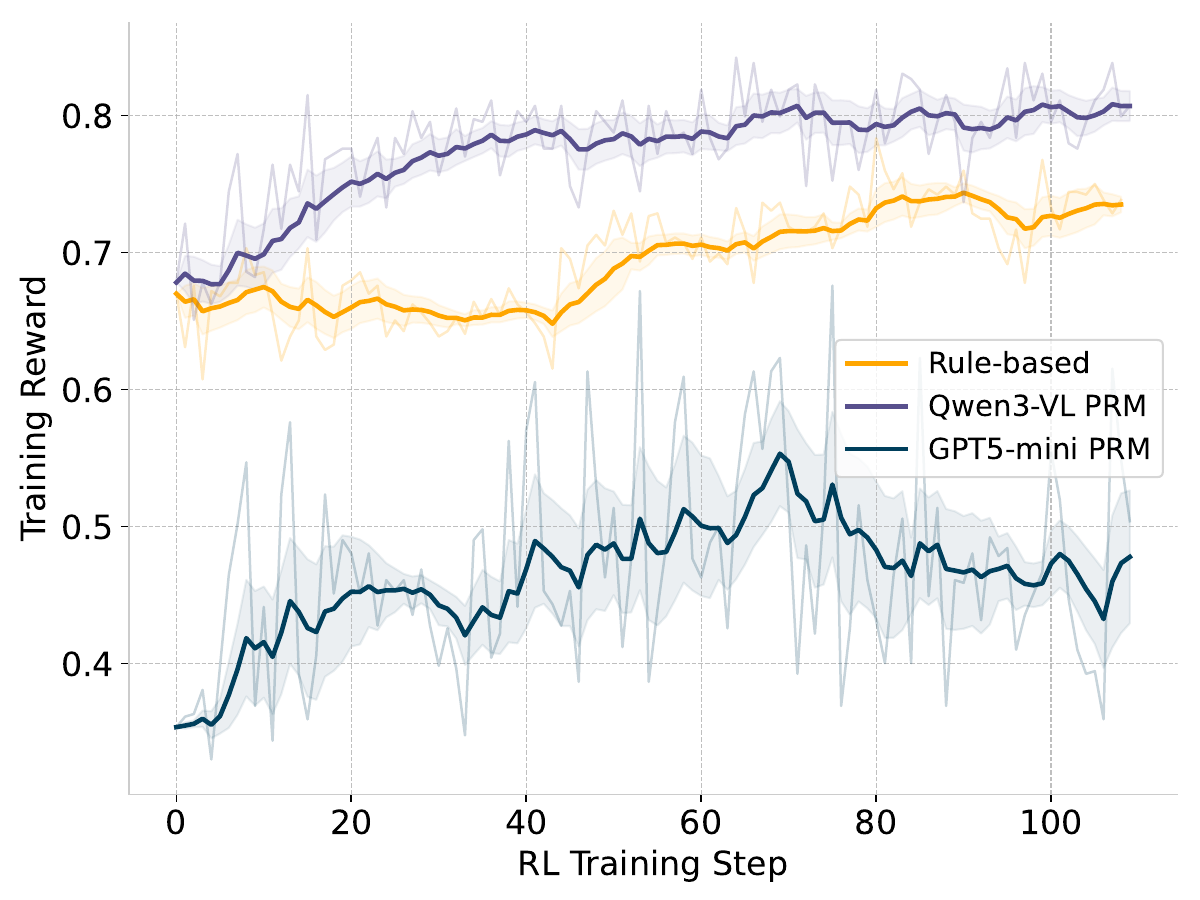}
    \caption{\textbf{Step-level rewards assigned during training} with moving average. \texttt{GPT-5-mini} assigns more conservative rewards, while \texttt{Qwen3-VL-4B} is more lenient on average. Despite this calibration gap, both PRMs achieve similar downstream policy performance, suggesting that GRPO is robust to differences in reward strictness through group normalization.}
    \label{fig:reward}
    \vspace{-0.3cm}
\end{figure}

\begin{table*}[t!]
\centering
\caption{\textbf{Task success rates on three online web benchmarks.} We compare PRO-CUA with open CUA models trained on large-scale expert or closed data, and with controlled baselines using Qwen3-VL backbones. PRO-CUA improves the base policy through iterative on-policy self-evolution without relying on expert demonstrations. \looseness=-1}
\label{tab:results}
\scalebox{0.75}{
\begin{tabular}{llcccc}
\toprule
\textbf{Training Paradigm} & \textbf{Method} & \textbf{Ext. Expert Steps} & \textbf{WebVoyager} & \textbf{Mind2Web-Live} & \textbf{OnlineMind2Web} \\
\midrule
\multirow{5}{*}{External expert / closed data}
& UI-TARS-1.5-7B & Closed data & 30.3 & 18.1 & 14.6 \\
& WebSTAR-7B & 100K & 47.0 & 17.0 & 17.0 \\
& WebSTAR-32B & 100K & 53.5 & 20.4 & 23.8 \\
& GUI-Libra-4B & 81K & -- & -- & 20.0 \\
& GUI-Libra-8B & 81K & -- & -- & 19.3 \\
\midrule
\multirow{4}{*}{Self-evolving 4B}
& Qwen3-VL-4B-Instruct & 0 & 27.5 & 18.1 & 16.7 \\
& FBC & 0 & 29.7 & 26.4 & 23.7 \\
& Rule-based Step-RL & 0 & 34.7 & 27.8 & \textbf{29.9} \\
& PRO-CUA & 0 & \textbf{42.4} & \textbf{34.7} & 28.8 \\
\midrule
\multirow{4}{*}{Self-evolving 8B}
& Qwen3-VL-8B-Instruct & 0 & 25.6 & 20.8 & 12.2 \\
& FBC & 0 & 31.8 & 23.6 & 26.9 \\
& Rule-based Step-RL & 0 & 33.8 & 25.0 & 26.2 \\
& PRO-CUA-8 & 0 & \textbf{43.2} & \textbf{30.6} & \textbf{28.2} \\
\bottomrule
\end{tabular}}
\end{table*}


Furthermore, we observe that the lightweight \texttt{Qwen3-VL-4B} PRM achieves similar downstream performance to \texttt{GPT-5-mini}. To better understand this result, we plot the moving average of step-level rewards assigned during optimization. As shown in \Cref{fig:reward}, the two PRMs exhibit substantially different reward calibration: \texttt{GPT-5-mini} is more conservative, with an average reward around 0.5, whereas \texttt{Qwen3-VL-4B} is more lenient, with an average reward above 0.7. One possible explanation is evaluator pedantry~\citep{zheng2023judging,li2024llms}, a phenomenon often observed in strong LLM-as-a-judge systems, where valid exploratory actions may be penalized for deviating from the evaluator's internal preference or stylistic prior. \looseness=-1

Importantly, despite calibration differences, both PRMs lead to comparable final policy performance. This suggests that PRO-CUA does not require perfectly calibrated absolute rewards. Since GRPO computes mean-centered advantages within each sampled group, binary PRM feedback only needs to provide useful local discrimination between acceptable and unacceptable candidate actions. The resulting updates aggregate these signals across many states, sampled candidates, and optimization steps, making the training process robust to differences in reward strictness across PRM backbones. \looseness=-1

\subsection{Main Results}
\paragraph{Benchmark evaluation}
\Cref{tab:results} reports the performance on three online web benchmarks. We include both representative open CUA models trained with large-scale expert or closed data, and controlled baselines initialized from the same Qwen3-VL backbones. Compared with prior open models, PRO-CUA is self-evolving: it improves the base policy through on-policy rollouts and PRM-guided step-level RL without relying on external expert demonstrations. Despite this weaker supervision assumption, PRO-CUA achieves competitive performance with expert-trained systems, while substantially improving over its own base models and controlled training baselines.

Within the controlled comparison, PRO-CUA achieves the strongest overall performance across both model sizes, with particularly large gains on WebVoyager and Mind2Web-Live. For the 4B model, PRO-CUA improves over FBC by 12.7\% on WebVoyager and 8.3\% on Mind2Web-Live in success rate, and improves over rule-based Step-RL by 7.7\% and 6.9\%, respectively. The gains remain consistent for the 8B model, where PRO-CUA outperforms rule-based Step-RL by 9.4\% on WebVoyager and 5.6\% on Mind2Web-Live. These results suggest that process-reward-guided step-level RL substantially improves the agent's ability to interact with dynamic web environments. The improvement is especially pronounced on WebVoyager, likely because its web-domain distribution is closest to our training queries.

Notably, PRO-CUA consistently improves over rule-based Step-RL despite both methods performing step-level reinforcement learning with GRPO. This suggests that the key advantage of PRO-CUA comes from the reward source rather than the optimization objective. Rule-based rewards provide reliable supervision only under the existence of golden answers, while our PRM grading is capable of evaluating intermediate steps across both successful and failed trajectories. We analyze this advantage in more detail below.

\begin{figure}[t!]
    \centering
    \includegraphics[width=0.8\linewidth]{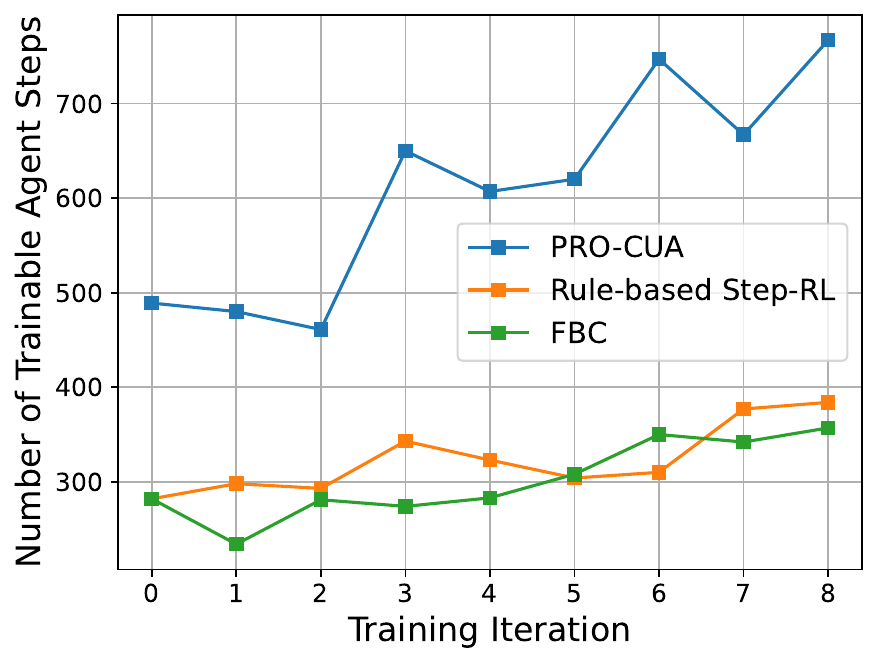}
    \caption{\textbf{Data utilization across training iterations.} PRO-CUA consistently yields more usable step-level training data than FBC and rule-based Step-RL because process rewards allow learning from both successful and failed finished trajectories, while the baselines rely on successful rollouts.}
    \label{fig:data_util}
    \vspace{-0.3cm}
\end{figure}

\paragraph{Data utilization} One key advantage of PRO-CUA is its significantly improved data utilization by learning from both successful and failed trajectories, bypassing the strict bottlenecks of only successful trajectories. To illustrate this, we further compare the data utilization of different training paradigms. In PRO-CUA, we apply a lightweight filtering strategy to remove redundant states. In particular, agents may occasionally get stuck on the same page for many steps, producing highly repetitive states and actions that provide little additional training signal. To reduce such redundancy, we retain only states from \textit{finished} trajectories, where the agent terminates the task and reports an answer within the maximum step budget. Importantly, finished trajectories are not necessarily successful: the final answer may still be incorrect.

\Cref{fig:data_util} shows the number of deployable step-level training examples at each iteration after filtering. PRO-CUA consistently yields substantially more deployable training steps than FBC and rule-based Step-RL. This is because both baselines rely on successful trajectories, while PRO-CUA can learn from both successful and failed trajectories by assigning process rewards to intermediate steps. This enables PRO-CUA to convert a larger fraction of on-policy interactions into trainable supervision, leading to better data utilization throughout iterative training. \looseness=-1

\section{Related Works}
\paragraph{Reinforcement learning for computer use agents} Filtered Behavior Cloning (FBC)~\citep{he2026webstarscalabledatasynthesis,xuaguvis,bai2026webgym} is the prevailing training paradigm for computer use agents. FBC first collects candidate trajectories from human annotators, stronger teacher models, or the agent itself, filters them by final task success, and then trains the policy with SFT on the retained successful trajectories. Recent work has explored RL for GUI and device-control agents to address the limitations of static imitation learning in dynamic environments. DigiRL~\citep{bai2024digirl} uses offline-to-online RL with advantage-weighted regression, where the actor is updated by imitating filtered high-advantage actions. Digi-Q~\citep{bai2025digiq} learns a Q-function from offline interaction data and extracts a policy by imitating the highest-scored candidate action. These methods go beyond standard behavior cloning, but they still rely on outcome-derived value estimates. For long-horizon GUI tasks, such outcome rewards are sparse and delayed, making it difficult to assign credit to individual intermediate actions.

This motivates recent work on step-level RL for GUI agents. GUI-R1~\citep{luo2025gui} and UI-R1~\citep{lu2026ui} apply R1-style rule-based RL for GUI action prediction, while GUI-Libra~\citep{yang2026gui} improves training under partially verifiable GUI rewards. However, the reward design of those methods still depends on verifiable rules or reference actions. Moreover, they typically construct training states from teacher-generated trajectories, which are often mismatched with the states encountered by the target policy. In contrast, PRO-CUA collects states on-policy from the current agent and uses visually grounded PRM feedback to score candidate thought-action pairs at those states. This preserves the credit-assignment benefit of step-level RL while reducing off-policy distribution shift and avoiding the rigidity of rule-based rewards, enabling learning from both successful and failed trajectories.


\paragraph{Process reward models} PRMs have achieved widespread success in reasoning-intensive domains such as mathematics~\citep{lightman2024let,wang2024math,zhang2025lessons} and code generation~\citep{dai2024process}. They are commonly used in two ways: \textit{test-time scaling}, where a PRM ranks sampled solutions or actions through Best-of-$N$ selection or reward-guided search~\citep{wang2024math,wang2025visualprm,xiong2025gui}, and \textit{training-time supervision}, where PRM scores serve as intermediate rewards for reinforcement learning~\citep{setlur2025rewarding,zhang2025reward}. Computer use tasks are structurally suited for step-level supervision because GUI navigation offers natively atomic steps (discrete clicks/typing), and a single erroneous action can derail the entire trajectory (e.g., closed tabs or failed verifications). However, test-time scaling is particularly costly in this setting: each candidate action may require environment interaction, screenshot processing, and PRM evaluation, making search-style methods substantially more expensive than single-policy execution.

Recent work has developed PRMs for web and computer-use agents. WebShepherd~\citep{chae2026webshepherd} trains a checklist-based PRM for web navigation, while WebArbiter~\citep{zhang2026webarbiter} improves it with principle-guided reasoning. However, these works primarily use PRMs as inference-time evaluators for action selection. This does not internalize the reward signal into the policy and is sensitive to PRM noise, since the agent directly executes the action preferred by the reward model. As CUARewardBench~\citep{lin2025cuarewardbench} suggests, even proprietary models remain primitive in step-level judging for CUA tasks, and most open-sourced PRMs are distilled from proprietary teachers rather than serving as reliable ground-truth evaluators. In contrast, PRO-CUA uses PRM as a source of training rewards in iterative on-policy RL. By aggregating noisy step-level signals across many states, sampled actions, and optimization updates, PRO-CUA is more robust than committing to a single PRM-selected action at inference time, as supported by our validation that \texttt{Qwen3-VL-4B} performs comparably to \texttt{GPT-5-mini}. Compared with offline filtering methods such as WebSTAR~\citep{he2026webstarscalabledatasynthesis}, PRO-CUA directly optimizes the policy with PRM scores, enabling both successful and failed on-policy trajectories to provide supervision.

\section{Conclusion}
We introduced PRO-CUA, an iterative step-level reinforcement learning framework for training computer use agents with process rewards. PRO-CUA decouples slow live environment interaction from policy optimization by first collecting on-policy states from the current agent and then optimizing the policy over PRM-graded candidate thought-action pairs. This design addresses two central bottlenecks in existing CUA training: the distribution shift caused by relying on teacher-generated states, and the limited data utilization caused by rule-based rewards that require successful trajectories or golden actions. By using flexible PRM feedback as step-level rewards, PRO-CUA can learn from both successful and failed finished trajectories and convert a larger fraction of on-policy interactions into trainable supervision. Experiments on three online web benchmarks show that PRO-CUA consistently improves over filtered behavior cloning and rule-based Step-RL, demonstrating the promise of process-reward-guided on-policy training for building more capable computer use agents.

\newpage
\section*{Limitations}
While PRO-CUA provides an effective framework for process-reward-guided training of computer use agents, several directions remain beyond the scope of this work.

First, we follow the standard CUA setup where the policy observes the task instruction, action history, and the current screenshot. We do not incorporate more advanced harnessing strategies such as long-term memory, retrieval, or explicit context engineering. These components are complementary to our training framework and could further improve performance on tasks that require long-range information tracking.

Second, our experiments only focus on web-based computer-use tasks. This setting provides a realistic and widely used testbed for CUA research because web environments are dynamic, visually diverse, and require long-horizon interaction. However, computer use agents are broader than web navigation alone, and many practical applications involve desktop software and mobile apps. These environments may introduce different interaction patterns, interface conventions, safety constraints, and state representations. While PRO-CUA is designed as a general training framework and does not rely on web-specific assumptions beyond the environment interface, validating its effectiveness beyond web benchmarks remains an important direction for future work.


\bibliography{custom}

\appendix

\section{Evaluation Details} \label{appendix:eval}
\begin{table*}[t!]
    \centering
    \caption{Websites with access problems in Mind2Web queries.}
    \scalebox{0.9}{
    \begin{tabular}{p{3cm}p{10cm}}
    \toprule
    Error     &  Websites \\
    \midrule
    Human verification & \texttt{bbb.org, redfin.com, extraspace.com, seatgeek.com, google.com/shopping, sec.gov, expedia.com, discogs.com, allrecipes.com} \\
    Access denied & \texttt{americashealthrankings.org, adoptapet.com, reddit.com, qatarairways.com, fedex.com, macys.com, gamestop.com, disney.com, kbb.com, accuweather.com} \\
    Website issue & \texttt{nba.com, target.com, phys.org, ups.com, kfc.com, cars.com, bestbuy.com, apartments.com, kohls.com} \\
    \bottomrule
    \end{tabular}}
    \label{tab:web_problem}
\end{table*}

In \Cref{tab:web_problem}, we detail specific websites that currently employ strict anti-bot mechanisms, preventing automated access. Consequently, we exclude all tasks requiring interaction with these domains from both our training and evaluation sets. It is important to note that these accessibility issues reflect the state of these platforms at the time of our experiments (May 2026); future updates may resolve these barriers and render the tasks viable again. Additionally, we omit tasks associated with Google, GitHub, and Allrecipes from the WebVoyager evaluation, as these platforms mandate rigorous human verification that falls outside the scope of our automated agent.

\section{Experimental Details} \label{appendix:exp}
For the WebVoyager results in \Cref{tab:results}, we directly from the original WebSTAR paper~\citep{he2026webstarscalabledatasynthesis}. Since Allrecipes and GitHub are deprecated as mentioned above, we compute the average performance for UI-TARS-1.5-7B, WebSTAR-7B and WebSTAR-32B without considering those domains to ensure a fair comparison.

Following the definition in Qwen3-VL~\citep{bai2025qwen3}, we present the action space of the computer use task in \Cref{tab:action_space}.
\begin{table}[t!]
\centering
\caption{Action space for computer use agents.}
\scalebox{0.7}{
\begin{tabular}{ll}
\toprule
\textbf{Action} & \textbf{Definition} \\
\midrule
\texttt{left\_click(x,y)} & Clicks at coordinates $(x,y)$ \\
\texttt{double\_click(x,y)} & Double-clicks at $(x,y)$ \\
\texttt{right\_click(x,y)} & Right-clicks at $(x,y)$ \\
\texttt{mouse\_move(x,y)} & Move the cursor to $(x,y)$ \\
\texttt{left\_click\_drag(x1,y1,x2,y2)} & Drags from $(x_1,y_1)$ to $(x_2,y_2)$ \\
\texttt{scroll(x,y,dir)} & Scrolls at $(x,y)$ in given direction \\
\texttt{type(content)} & Types text \\
\texttt{hotkey(keys)} & Presses hotkey \\
\texttt{wait()} & Pauses 5s \\
\texttt{goback()} & Go back to the previous page \\
\texttt{finished(content)} & Complete task with final answer \\
\bottomrule
\end{tabular}}
\label{tab:action_space}
\end{table}

\section{Implementation Details of Rule-Based Rewards}
\label{appendix:rule_based_reward}

For the rule-based Step-RL baseline, we implement an automated verifier following GUI-Libra~\citep{yang2026gui}. Each sampled rollout produces a structured prediction string $y$ following the required output format:
\begin{align*}
y = \texttt{<think>} \cdots \texttt{</think>}\texttt{<answer>} a \texttt{</answer>},
\end{align*}
where the \texttt{<answer>} block contains a structured action $a$ that can be parsed into a JSON object: $a = \{\texttt{action\_type},\texttt{description},\texttt{value},\texttt{point\_2d}\}.$
Here, \texttt{action\_type} specifies the primitive GUI operation, \texttt{value} denotes the text input when applicable, and $\texttt{point\_2d} \in \mathbb{R}^2$ denotes the target screen coordinate, or \texttt{none} when the action does not require grounding.

We define the rule-based reward as a weighted combination of a format reward and an action-correctness reward:
\begin{align*}
\tilde r(s,a)
=
w_{\mathrm{fmt}} r_{\mathrm{fmt}}
+
(1-w_{\mathrm{fmt}}) r_{\mathrm{acc}},
\end{align*}
where $w_{\mathrm{fmt}} \in [0,1]$. In our experiments, we set $w_{\mathrm{fmt}}=0.1$, so the reward primarily reflects action correctness while still encouraging valid structured outputs.

\paragraph{Format reward}
The format reward $r_{\mathrm{fmt}}$ checks whether the model output can be parsed correctly. We set $r_{\mathrm{fmt}}=1$ if the output contains valid \texttt{<think>} and \texttt{<answer>} tags and the \texttt{<answer>} block can be parsed into the required JSON schema. Otherwise, $r_{\mathrm{fmt}}=0$.

\paragraph{Accuracy reward}
The accuracy reward $r_{\mathrm{acc}}$ evaluates whether the predicted action matches the reference action from a successful trajectory. We decompose it into three components:
\begin{align*}
r_{\mathrm{acc}}
=
r_{\mathrm{type}} \cdot r_{\mathrm{value}} \cdot r_{\mathrm{ground}},
\end{align*}
where $r_{\mathrm{type}}$ evaluates the action type, $r_{\mathrm{value}}$ evaluates the textual input, and $r_{\mathrm{ground}}$ evaluates coordinate grounding.

First, the action-type reward checks whether the predicted action type matches the reference action type:
\begin{align*}
r_{\mathrm{type}}
=
\mathbbm{1}\left[
\texttt{action\_type}(a)
=
\texttt{action\_type}(a^\star)
\right],
\end{align*}
where $a^\star$ denotes the reference action.

Second, for actions involving text input, the value reward compares the predicted value $v$ with the reference value $v^\star$ using word-level F1:
\begin{align*}
r_{\mathrm{value}}
=
\mathbbm{1}\left[
\mathrm{F1}(v,v^\star) > 0.5
\right].
\end{align*}
For actions that do not require a textual value, we set $r_{\mathrm{value}}=1$.

Third, the grounding reward evaluates whether the predicted coordinate $\mathbf{u}$ falls inside the reference bounding box $b^\star$:
\begin{align*}
r_{\mathrm{ground}}
=
\mathbbm{1}\left[
\mathbf{u} \in b^\star
\right].
\end{align*}
For actions that do not require coordinate grounding, we set $r_{\mathrm{ground}}=1$.

This rule-based reward provides reliable supervision when a reference action is available. However, it is inherently tied to successful trajectories with golden actions. It may assign low rewards to valid alternative actions that deviate from the reference, and cannot be applied to states from failed trajectories where no verified reference action is available. This limitation motivates our use of PRM-based rewards in PRO-CUA.

\section{Potential Risks}
This paper presents work whose goal is to advance the field of NLP. There are many potential societal consequences of our work, none which we feel must be specifically highlighted here.

\section{Licenses}
Both WebVoyager and OpenWebVoyager are under Apache-2.0 license. Both Online-Mind2Web Mind2Web-Live are under cc-by-4.0 license.

\section{Prompt for PRM Grading} \label{appendix:prompt}
\begin{tcolorbox}[guiPrompt, title={PRM grading prompt},
  float*=h,            
  floatplacement=tbp,  
  width=\textwidth
]
\small
You are an expert evaluator grading a Computer-Use Agent. Your role is to evaluate whether the agent’s proposed next action is the strictly correct and necessary step to advance the given task.

You are provided with:

1. The overarching task instruction.

2. The history of actions taken so far.

3. The CURRENT screenshot (the state immediately BEFORE the proposed action), annotated to show the proposed target of the action.

4. The proposed Thought and Action Code.

The screenshot is an annotated visualization of the proposed action, not a raw screenshot:

- Red marks, arrows, or points indicate where the proposed action is targeting.

- Small index labels and overlay text are part of the annotation.

- Use these annotations to judge whether the proposed action is correctly grounded on the UI.

- Do not confuse the annotation itself with a native page element.

<task\_instruction>
{instruction}
</task\_instruction>

<history\_actions>
{history\_actions}
</history\_actions>

<proposed\_action>
Step {step\_index}: {action\_code}
</proposed\_action>

Evaluation Criteria
You must evaluate the proposed action and output a binary decision: is the action CORRECT or INCORRECT?

An action is INCORRECT if it exhibits ANY of the following flaws:

- Grounding Failure: The code targets the wrong coordinates, a non-existent element, or the wrong input field based on the provided screenshot.

- Hallucination: The agent assumes a state that is not visually present.

- Inefficiency/Redundancy: The action needlessly repeats a past step from the history, performs useless scrolling, or wastes a step without advancing the task.

- Logical Progression Failure: The action executes successfully but does not move the agent closer to the final goal.

An action is CORRECT ONLY if it is visually grounded, mathematically accurate, and actively advances the task toward completion.

Output Format
Provide a rigorous step-by-step reflection. You must perform a "mental rollout" to predict the consequences of the action before determining if it facilitates task completion. Think about other alternatives that might result in better outcome than the proposed action, and if there exists such alternative with strictly better outcome, make the action as incorrect. Then, output a strictly valid JSON block.

<analysis\_process>

1. [Current State Assessment]: What is currently visible on the screen? What is the immediate blocker to completing the task?

2. [Target Verification]: Does the proposed code correctly and accurately target the intended UI element in the screenshot?

3. [Mental Rollout]: If this exact code is executed, what will happen? (e.g., "A dropdown menu will appear," "The page will scroll down," "The text 'shoes' will be typed").

4. [Task Alignment]: Does this predicted outcome meaningfully and efficiently advance the task? Or is it a redundant/wasteful action given the history?

5. [Final Verdict]: Conclude whether the step is Correct or Incorrect.

</analysis\_process>

```json

{

  "is\_correct": boolean,
  
  "reflection": "A 1-2 sentence summary of why the action was marked correct or incorrect."
  
}"""

\end{tcolorbox}

\end{document}